\documentclass{article} 
\usepackage{iclr2022_conference,times}

\usepackage{amsmath,amsfonts,bm}

\def\eqref#1{equation~\ref{#1}}

\def\1{\bm{1}}

\DeclareMathAlphabet{\mathsfit}{\encodingdefault}{\sfdefault}{m}{sl}
\SetMathAlphabet{\mathsfit}{bold}{\encodingdefault}{\sfdefault}{bx}{n}

\newcommand{\E}{\mathbb{E}}

\usepackage{hyperref}
\usepackage{url}
\usepackage[utf8]{inputenc}
\usepackage{amsmath}
\usepackage{amsfonts}
\usepackage{amssymb}
\usepackage{tabularx}
\usepackage{booktabs}
\usepackage{mathtools}
\usepackage{algpseudocode}
\DeclareMathOperator{\U}{\mathbf{U}}
\DeclareMathOperator{\DKL}{\text{D}_\text{KL}}

\usepackage{color, colortbl}
\definecolor{Gray}{gray}{0.95}
\title{Mixture-of-Variational-Experts for Continual Learning}

\author{Heinke Hihn\thanks{Corresponding Author \newline Code is available at \url{https://sites.google.com/view/hvcl/home} \newline Preprint}~ \& Daniel A. Braun \\
Institute of Neural Information Processing\\
Ulm University, Ulm, Germany \\
\texttt{\{heinke.hihn,daniel.braun\}@uni-ulm.de} \\
}

\iclrfinalcopy 
\begin{document}

\maketitle

\begin{abstract}
One weakness of machine learning algorithms is the poor ability of models to solve new problems without forgetting previously acquired knowledge. The Continual Learning (CL) paradigm has emerged as a protocol to systematically investigate settings where the model sequentially observes samples generated by a series of tasks. In this work, we take a task-agnostic view of continual learning and develop a hierarchical information-theoretic optimality principle that facilitates a trade-off between learning and forgetting. We discuss this principle from a Bayesian perspective and show its connections to previous approaches to CL. Based on this principle, we propose a neural network layer, called the Mixture-of-Variational-Experts layer, that alleviates forgetting by creating a set of information processing paths through the network which is governed by a gating policy. Due to the general formulation based on generic utility functions, we can apply this optimality principle to a large variety of learning problems, including supervised learning, reinforcement learning, and generative modeling. We demonstrate the competitive performance of our method in continual supervised learning and in continual reinforcement learning.
\end{abstract}

\section{Introduction}
Acquiring new skills without forgetting previously acquired knowledge is a hallmark of human and animal intelligence. Biological learning systems leverage task-relevant knowledge from preceding learning episodes to guide subsequent learning of new tasks to accomplish this. Artificial learning systems, such as neural networks, usually lack this crucial property and experience a problem coined "catastrophic forgetting" \citep{mccloskey1989catastrophic}. Catastrophic forgetting occurs when we naively apply machine learning algorithms to solve a sequence of tasks $T_{1:t}$, where adaptation to task $T_t$ prompts overwriting of parameters learned for tasks $T_{1:t-1}$.

The Continual Learning (CL) paradigm \citep{thrun1998lifelong} has emerged as a way to investigate such problems systematically. We can divide CL approaches into three broad categories: rehearsal and memory consolidation, regularization and weight consolidation, and architecture and expansion methods. Rehearsal methods train a generative model to learn the data-generating distribution to reproduce data of old tasks \citep{shin2017continual,rebuffi2017icarl}. In contrast, regularization methods \citep[e.g.,][]{kirkpatrick2017overcoming,ahn2019uncertainty,han2021continual} introduce an additional constraint to the learning objective to prevent changes in task-relevant parameters. Finally, CL can be achieved by modifying the design of a model during learning \citep[e.g.,][]{lin2019conditional,rusu2016progressive,golkar2019continual}. 

Despite recent progress in CL, there are still open questions \citep{parisi2019continual}. For example, most existing algorithms share a significant drawback in that they require task-specific knowledge, such as the number of tasks and which task is currently at hand. Approaches sharing this drawback are multi-head methods \citep[e.g.,][]{khatib2019strategies,nguyen2017variational,ahn2019uncertainty} and methods that  compute a per-task loss, which requires storing old weights and task information.\citep[e.g.,][]{kirkpatrick2017overcoming,zenke2017continual,sokar2021self,yoon2018lifelong,chaudhry2021using,han2021contrastive}. Extracting relevant task information is a difficult problem, in particular when distinguishing tasks without contextual input \citep{hihn2020specialization}. Thus, providing the model with such task-relevant information yields overly optimistic results \citep{chaudhry2018riemannian}. 

To deal with more realistic CL scenarios, therefore, models must learn to compensate for the lack of auxiliary information. The approach we propose here tackles this problem by formulating a hierarchical learning system, that allows us to learn a set of sub-modules specialized in solving particular tasks. To this end, we introduce hierarchical variational continual learning (HCVL) and devise the mixture-of-variational-experts layer (MoVE layers) as an instantiation of HVCL. MoVE layers consist of $M$ experts governed by a gating policy, where each expert maintains a posterior distribution over its parameters alongside a corresponding prior. A sparse selection reduces computation as only a small subset of the parameters must be updated during the back-propagation of the loss \citep{Shazeer2017}. To mitigate catastrophic forgetting we condition the prior distributions on previously observed tasks and add a penalty term on the Kullback-Leibler-Divergence between the expert posterior and its prior.

In ensemble methods two main questions arise. The first one concerns the question of optimally selecting ensemble members using appropriate selection and fusion strategies \citep{Kuncheva2004}. The second one, is the question of how to ensure expert diversity \citep{kuncheva2003measures,bian2021when}. We argue that ensemble diversity benefits continual learning and investigate two complementary diversity objectives: the entropy of the expert selection process and a similarity measure between different experts based on Wasserstein exponential kernels in the context of determinantal point processes \citep{kulesza2012determinantal}.

To summarize, our contributions are the following: $(i)$ we extend VCL to a hierarchical multi-prior setting, $(ii)$ we derive a computationally efficient method for task-agnostic continual learning from this general formulation, $(iii)$ to improve expert specialization and diversity, we introduce and evaluate novel diversity measures.

This paper is structured as follows: we introduce our method in Section \ref{sec:clmove}, we design, perform, and evaluate the main experiments in Section \ref{sec:experiments}, in Section~\ref{sec:discussion}, we discuss novel aspects of the current work in the context of previous literature and conclude with a final summary in Section \ref{sec:conclusion}.

\section{Hierarchical Variational Continual Learning}
\label{sec:clmove}
\label{sec:hvcl}
In this section we first extend the variational continual learning (VCL) setting introduced by \cite{nguyen2017variational} to a hierarchical multi-prior setting and then introduce a neural network implementation as a generalized application of this paradigm in Section~\ref{sec:sparse}.

VCL describes a general learning paradigm wherein an agent stays close to an old strategy ("prior") it has learned on a previous task $t-1$ while learning to solve a new task $t$ ("posterior"). 
Given datasets of input-output pairs $\mathcal{D}_t = \{x_t^i, y_t^i\} _{i=0}^{N_t}$  of tasks $t \in \{1, ..., T\}$,  the main learning objective of minimizing the log-likelihood $\log p_\theta(y_t^{i}\vert x_t^{i})$ for task $t$ is augmented with an additional loss term in the following way:
\begin{equation}
\label{eq:vcl}
\begin{aligned}
\mathcal{L}_{\text{VCL}}^t = \sum_{i=1}^{N_t} \mathbb{E}_{\theta}\left[\log p_\theta(y_t^{i}\vert x_t^{i})\right] - \DKL\left[p_t(\theta)\vert \vert p_{t-1}(\theta) \right],
\end{aligned}
\end{equation}
where $p(\theta)$ is a distribution over the models parameter $\theta$ and $N_t$ is the number of samples for task $t$.
The constraint encourages the agent to find an optimal trade-off between solving a new task and retaining knowledge about old tasks.  Over the course of $T$ datasets, Bayes' rule then recovers the posterior
\begin{equation}
\label{eq:bayesinference}
p(\theta\vert \mathcal{D}_{1:T}) \propto p(\theta\vert \mathcal{D}_{1:T-1})p(\mathcal{D}_T\vert \theta)
\end{equation}
which forms a recursion: the posterior after seeing $T$ datasets is obtained by multiplying the posterior after $T-1$ with the likelihood and normalizing accordingly.

This multi-head strategy has two main drawbacks: $(i)$ it introduces an organizational overhead due to the growing number of network heads, and $(ii)$ task boundaries must be known at all times, making it unsuitable for more complex continual learning settings. In the following we argue that we can alleviate these problems by combining multiple decision-makers with a learned selection policy.

To extend VCL to the hierarchical case, we  assume that samples are drawn from a set of $M$ independent data generating processes, i.e.,\ the likelihood is given by a mixture model $p(y\vert x) = \sum_{m=1}^M p(m\vert x)p(y\vert m,x)$. We define an indicator variable $z \in Z$, where $z_m^{i,t}$ is $1$ if the output $y_i^t$ of sample $i$ from task $t$ was generated by expert $m$ and zero otherwise:
\begin{equation}
\label{eq:mixturelikelihood}
p(y_t^i\vert x_t^i,\Theta) = \sum_{m=1}^M p(z_t^{i,m}\vert x_t^i,\vartheta)p(y_t^i\vert x_t^i,\omega_m),
\end{equation}
where $\vartheta$ are the parameters of the selection policy, $\omega_m$ the parameters of the $m$-th expert, and $\Theta = \{\vartheta, \{\omega_m\}_{m=1}^M\}$ the combined model parameters. The posterior after observing $T$ tasks is then given by
\begin{align}
\label{eq:hierarchicalbayesinference}
p(\Theta\vert \mathcal{D}_{1:T}) & \propto p(\vartheta)p(\omega) \prod_{t=1}^{T}\prod_{i=1}^{N_t} \sum_{m=1}^M p(z_t^{i,m}\vert x_t^i,\vartheta)p(y_t^i\vert x_t^i,\omega_m) \\ 
& =  p(\Theta)\prod_{t=1}^T p(\mathcal{D}_t\vert \Theta) \propto p(\Theta\vert \mathcal{D}_{1:T-1})p(\mathcal{D}_T\vert \Theta). \nonumber
\end{align}
The Bayes posterior of an expert $p(\omega_m\vert \mathcal{D}_{1:T})$ is recovered by computing the marginal over the selection variables $Z$. Again, this forms a recursion, in which the posterior $p(\Theta\vert \mathcal{D}_{1:T})$ depends on the posterior after seeing $T-1$ tasks and the likelihood $p(\mathcal{D}_T\vert \Theta)$. Finally, we formulate the HVCL objective for task $t$ as:
\begin{equation}
\label{eq:hvcl}
\begin{aligned}
\mathcal{L}_{\text{HVCL}}^t & = \sum_{i=1}^{N_t} \mathbb{E}_{p(\Theta)}\left[\log p(y_t^{i}\vert x_t^{i}, \Theta)\right] - \DKL\left[p_t(\vartheta)\vert \vert p_{1:t-1}(\vartheta)\right] - \DKL\left[p_t(\omega)\vert \vert p_{1:t-1}(\omega)\right],
\end{aligned}
\end{equation}
where $N_t$ is the number of samples in task $t$, and the likelihood $p(y,x\vert \Theta)$ is defined as in~\eqref{eq:mixturelikelihood}. 

\subsection{Sparsely Gated Mixture-of-Variational Layers}
\label{sec:sparse}

\begin{figure*}[t!]
\centering
\includegraphics[width=0.95\textwidth, trim={16cm 0.5cm 3.5cm .5cm}, clip]{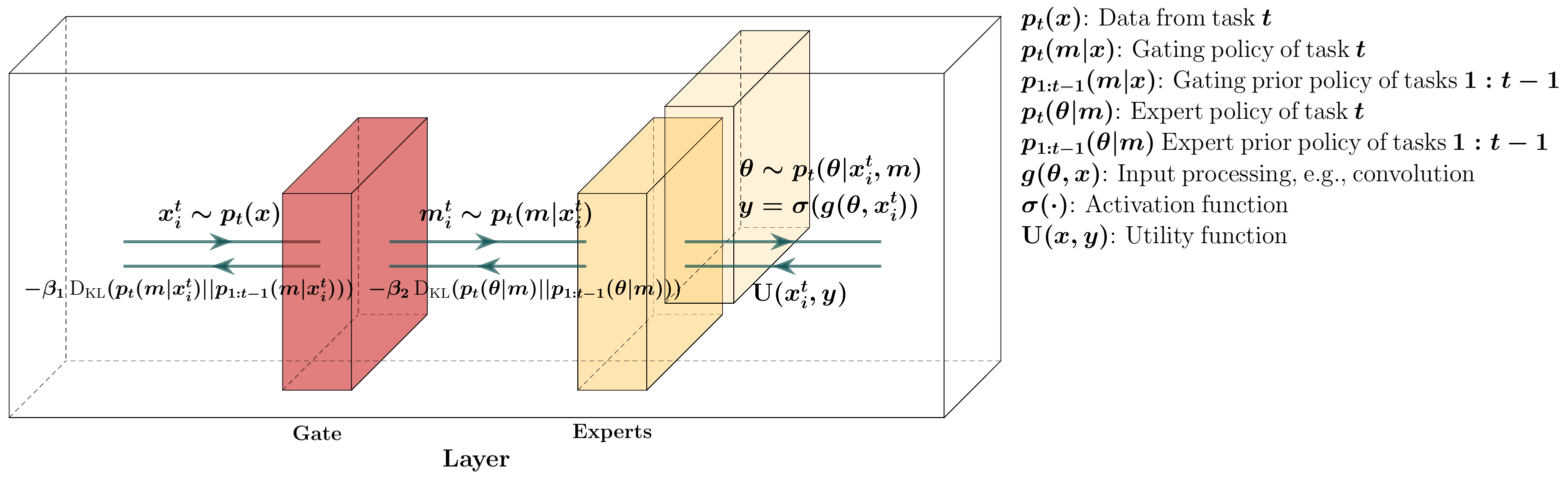}
\caption{This figure illustrates our proposed design. Each layer implements a top-$k$ expert selection conditioned on the output of the previous layer. Each expert $m$ maintains a distribution over its weights $p(\theta\vert m) = \mathcal{N}(\mu_m, \sigma_m)$ and a set of bias variables $b_{m}$.}
\label{fig:architecture}
\end{figure*}

As we aim to tackle not only supervised learning problems, but also reinforcement learning problems, we assume in the following a generic scalar utility function $\U(x,f_\theta(x))$ that depends both on the input $x$ and the parameterized agent function
$f_\theta(x)$ that generates the agent's output $y$. We assume the agents output function $f_\theta(x)$ is composed of multiple layers.
Our layer design builds on the sparsely gated Mixture-of-Expert (MoE) layers \citep{Shazeer2017}, which in turn draws on the paradigm introduced by \cite{Jacobs1991}. MoEs consist of a set of $m$ experts $M$ and a gating network $p(m\vert x)$ whose output is a (sparse) $m$-dimensional vector. All experts have an identical architecture but separate parameters. Let $p(m\vert x)$ be the gating output and $p(y\vert m,x)$ the response of an expert $m$ given input $x$. The layer's output is then given by a weighted sum of the experts responses. To save computation time we employ a top-$k$ gating scheme, where only the $k$ experts with highest gating activation are evaluated and use an additional penalty that encourages gating sparsity (see Section \ref{sec:sparse}). In all our experiments we set $k = 1$, to drive expert specialization (see Section \ref{sec:specdiv}) and reduce computation time.
We implement the learning objective for task $t$ as layer-wise regularization in the following way:
\begin{equation}
\begin{aligned}
\mathcal{L}_{\text{MoVE}}^{t} = \sum_{i=1}^{N_t} \bigg[\mathbb{E}_{\Theta}\left[\U(x_i^t, f_\Theta(x_i^t))\right] - & \sum_{l=1}^{L} \mathbb{E}_{p_t^l(m\vert x_i^t), p_t^l(\theta\vert m)}\bigg[\beta_1 \DKL\left[p_t^l(m\vert x_i^t)\vert \vert p_{1:t-1}^l(m\vert x_i^t)\right] \\
&  + \beta_2 \DKL\left[p_t^l(\theta\vert m)\vert \vert p_{1:t-1}^l(\theta\vert m) \right] \bigg] \bigg],
\end{aligned}
\label{eq:move}
\end{equation}
where $L$ is the total number of layers,  $\Theta = \{\theta, \{\vartheta_m\}_{m=1}^M\}$ the combined parameters,and the temperature parameters $\beta_{1,2}$ govern the layer-wise trade-off between utility and information-cost. 

Thus, we allow for two major generalizations compared to \eqref{eq:hvcl}: in lieu of the log-likelihood we allow for generic utility functions, and instead of applying the constraint on the gating parameters, we apply it directly on the gating output distribution $p(m\vert x)$. This implies, that the weights of the gating policy are not sampled. Otherwise the gating mechanism would involve two stochastic steps: one in sampling the weights and a second one in sampling the expert index. This potentially high selection variance hinders expert specialization (see Section \ref{sec:specdiv}). Encouraging the gating policy to stay close to its prior also ensures that similar inputs are assigned to the same expert. Next we consider how we could extend objective \ref{eq:move} further by additional terms that encourage diversity between different experts.

\subsection{Encouraging Expert Diversity}
\label{subsec:diversity}
\label{sec:paramdiv}
In the following, we argue that a diverse set of experts may mitigate catastrophic forgetting in continual learning, as experts specialize more easily in different tasks, which improves expert selection. We present two expert diversity objectives. The first one arises directly from the main learning objective and is designed to act as a regularizer on the gating policy while the second one is a more sophisticated approach that aims for diversity in the expert parameter space. The latter formulation introduces a new class of diversity measures, as we discuss in more detail in Section \ref{sec:discussion}.

\subsubsection{Diversity through Specialization}
\label{sec:specdiv}
\begin{figure}[t!]
\centering
\includegraphics[width=0.4\textwidth, trim={.25cm 27.75cm 15.5cm 0cm}, clip]{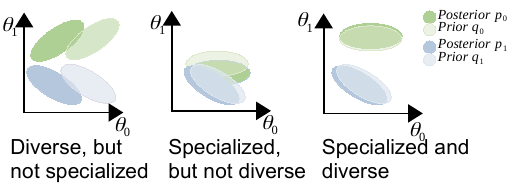}
\includegraphics[width=0.5\textwidth, trim={0cm 27cm 11.75cm 0cm}, clip]{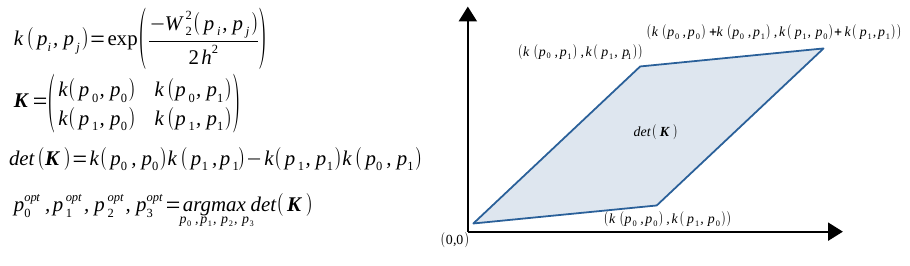}
\caption{Left: We seek experts that are both specialized, i.e., their posterior $p$ is close to their prior $q$, and diverse, i.e., posteriors are sufficiently distant from one another. Right: To this effect, we maximize the determinant of the kernel matrix $K$, effectively filling the feature space. In the case of two experts this would mean to maximize $det(\textbf{K}) = 1 -  k(p_0,p_1)$, which we can achieve by maximizing the Wasserstein-2 distance between the posteriors.}
\label{fig:specialization_diversity}
\end{figure}
The relationship between objectives of the form described by~\eqref{eq:move} with the emergence of expert specialization has been previously investigated for simple learning problems \citep{Genewein2015} and in the context of meta-learning \citep{hihn2020specialization}, but not in the context of continual learning. We assume a two-level hierarchical system of specialized decision-makers where low-level decision-makers $p(m\vert x)$ select which high-level decision-maker $p(y\vert m,x)$ serve as experts for a particular input $x$. By co-optimizing 
\begin{equation}
\label{eq:par_mutual}
\max_{p(y\vert x,m), p(m\vert x)} \E[\mathbf{U}(x,y)] - \beta_1 I(X;M) - \beta_2 I(X;Y\vert M),
\end{equation}
the combined system finds an optimal partitioning of the input space $X$, where $I(\cdot\vert \cdot)$ denotes the (conditional) mutual information between random variables. 
In fact, the hierarchical VCL objective given by \eqref{eq:hvcl} can be regarded as a special case of the information-theoretic objective given by \eqref{eq:par_mutual}, if we interpret the prior as the learning strategy of task $t-1$ and the posterior as the strategy of task $t$, and set $\beta = 1$. All these hierarchical decision systems correspond to a multi prior setting, where different priors associated with different experts can specialize on different sub-regions of the \emph{input space}. In contrast, specialization in the context of continual learning can be regarded as the ability of partitioning the \emph{task space}, where each expert decision-maker $m$ solves a subset of old tasks $T^m \subseteq T_{1:t}$. In both cases, expert diversity is a natural consequence of specialization if the gating policy $p(m\vert x)$ partitions between the experts. 

In addition to the implicit pressures for specialization already implied by~\eqref{eq:move}, here we investigate the effect of an additional entropy cost. Inspired by recent entropy regularization techniques \citep{eysenbach2018diversity,galashov2019information,grau2018soft}, we aim to improve the gating policy by introducing the entropy cost
\begin{equation}
\frac{\beta_3}{N_b}\sum_{x \in \mathcal{B}} H(M\vert X) - H(M),
\end{equation}
where $M$ is the set of experts and $X$ the inputs in a mini-batch $\mathcal{B}$ of size $N_b$, and $\beta_3$ a weight. By maximizing the conditional entropy $H(M\vert X)$ we encourage high certainty in the expert gating and by minimizing the marginal entropy $H(M)$ we prefer solutions that minimize the number of active experts. We compute these values batch-wise, as the full entropies over $p(x)$ are not tractable.

\subsubsection{Parameter Diversity}
Our second diversity formulation builds on Determinantal Point Processes (DPPs) \citep{kulesza2012determinantal}, a mechanism that produces diverse subsets by sampling proportionally to the determinant of the kernel matrix of points within the subset \citep{macchi1975coincidence}. A point process $P$ on a ground set $Y$ is a probability measure over finite subsets of $Y$. $P$ is a DPP if, when $Y$ is a random subset drawn according to $P$, we have, for every$A \subset Y$, $P(A \subset Y) = det(K_A)$ for some real, symmetric $N\times N$ matrix $K$ indexed by the elements of $Y$. Here, $K_A = [K_{ij}]_{i,j\in A}$ denotes the restriction of $K$ to the entries indexed by elements of $A$, and we adopt $det(K_\emptyset) = 1$.  If $A = \{i\}$ is a singleton, then we have $P(i\in Y) = K_{i,i}$. In this case, the diagonal of $K$ gives the marginal probabilities of inclusion for individual elements of $Y$. Since $P$ is a probability measure, all principal minors of $K$ must be nonnegative, and thus $K$ itself must be positive semidefinite, which we can achieve by constructing $K$ by a kernel function $k(x_0,x_1)$, such that for any $x_0,x_1 \in \mathcal{X}$:
\begin{equation}
\label{eq:kernel}
k(x_0,x_1) = \langle\phi(x_0),\phi(x_1)\rangle_\mathcal{F},
\end{equation}
where $\mathcal{X}$ is a vector space and $\mathcal{F}$ is a inner-product space such that $\forall x \in \mathcal{X}: \phi(x) \in \mathcal{F}$. 
Specifically, we use a exponential kernel based on the Wasserstein-2 distance $W(p,q)$ between two probability distributions $p$ and $q$. The $p^\text{th}$ Wasserstein distance between two probability measures $p$ and $q$ in $P_p(M)$ is defined as
\begin{equation}
\label{eq:wasserstein}
W_p (p, q):=\left( \inf_{\gamma \in \Gamma (p, q)} \int_{M \times M} d(x, y)^p \, \mathrm{d} \gamma (x, y) \right)^{1/p},
\end{equation}
where $\Gamma(p,q)$ denotes the collection of all measures on $M \times M$ with marginals $p$ and $q$ on the first and second factors. Let $p$ and $q$ be two isotropic Gaussian distributions and $W_2^2(q,p)$ the Wasserstein-2 distance between $p$ and $q$. The exponential Wasserstein-2 kernel is then defined by 
\begin{equation}
\label{eq:w2expkernel}
k(p,q) = \exp\left(-\frac{W_2^2(p,q)}{2h^2}\right),
\end{equation}
where $h$ is the kernel width. We show in Appendix \ref{app:wasserkernel} that \eqref{eq:w2expkernel} gives a valid kernel. This formulation has two properties that make it suitable for our purpose. Firstly, the Wasserstein distance is symmetric, i.e., $W_2^2(p,q) = W_2^2(q,p)$, which in turn will lead to a symmetric kernel matrix. This is not true for other similarity measures on probability distributions, such as $\DKL$ \citep{cover2012elements}. Secondly, if $p$ and $q$ are Gaussian and mean-field approximations, i.e., covariance matrices are given by diagonal matrices, i.e., $\Sigma_p = \text{diag}(d_p)$ and $\Sigma_q = \text{diag}(d_q)$, $W_2^2(p,q)$ can be computed in closed form as 
\begin{equation}
\label{eq:gaussianwasser}
W_2^2(p,q) = \vert \vert \mu_p - \mu_q\vert \vert _2^2 + \vert \vert \sqrt{d_p} - \sqrt{d_q}\vert \vert _2,
\end{equation}
where $\mu_{p,q}$ are the means and $d_{p,q}$ the diagonal entries of distributions $p$ and $q$. We provide a more detailed derivation of \eqref{eq:gaussianwasser} in Appendix \ref{app:wassergaussian}. From a geometric perspective, the determinant of the kernel matrix represents the volume of a parallelepiped spanned by feature maps corresponding to the kernel choice. We seek to maximize this volume, effectively filling the parameter space -- see Figure \ref{fig:specialization_diversity} for an illustration.

\section{Experiments}
\label{sec:experiments}
\begin{table}[t!]
\setlength{\tabcolsep}{2pt} 
\begin{minipage}{0.49\columnwidth}
\centering
\tiny
\begin{tabularx}{0.99\columnwidth}{*{3}l}
\toprule
\textbf{Baselines} &  \textbf{S-MNIST} & \textbf{P-MNIST} \\
\midrule
\rowcolor{Gray}
Dense Neural Network &  86.15 ($\pm$1.00)  & 17.26 ($\pm$0.19) \\
Offline re-training + task oracle &  99.64 ($\pm$0.03)& 97.59 ($\pm$0.02) \\
\midrule
\textbf{Single-Head and Task-Agnostic} & & \\
\midrule
\rowcolor{Gray}
HVCL (ours) & 97.50 ($\pm$0.33) & 97.07 ($\pm$0.62) \\
HVCL w/ GR (ours) & 98.60 ($\pm$0.35) & 97.47 ($\pm$0.52) \\
\rowcolor{Gray}
UGCL w/ BNN \citep{ebrahimi2020uncertainty} & 97.70 ($\pm$0.03) & 92.50 ($\pm$0.01) \\
Brain-inspired RtF \citep{van2020brain} & 99.66 ($\pm$0.13) & 97.31 ($\pm$0.04) \\
\rowcolor{Gray}
HIBNN \citep{kessler2021hierarchical} & 91.00 ($\pm$2.20) & 93.70 ($\pm$0.60)\\
BCL \citep{raghavan2021formalizing} & 98.71 ($\pm$0.06) & 97.51 ($\pm$0.05) \\
\rowcolor{Gray}
TLR \citep{mazur2021target} & 80.64 ($\pm$1.25) &\\
\midrule
\textbf{Multi-Head and Task-Aware} & & \\
\midrule
\rowcolor{Gray}
DGR+distill. \citep{shin2017continual} & 99.59 ($\pm$0.40) & 97.51 ($\pm$0.04) \\
VCL \citep{nguyen2017variational} & 98.50 ($\pm$1.78) & 96.60 ($\pm$1.34) \\
\rowcolor{Gray}
CURL \citep{rao2019continual} & 99.10 ($\pm$0.06)& \\
SAML \citep{sokar2021self} & 97.95 ($\pm$0.07) & \\
\rowcolor{Gray}
DEN \citep{yoon2018lifelong} & 99.26 ($\pm$0.01) & \\
\bottomrule
\end{tabularx}
\end{minipage}
\hfill
\begin{minipage}{0.49\columnwidth}
\tiny
\centering
\begin{tabularx}{0.99\columnwidth}{*{3}l}
\toprule
\textbf{Baselines} &  \textbf{Split-CIFAR-10} & \textbf{CIFAR-100} \\
\midrule
\rowcolor{Gray}
Conv. Neural Network &  66.62 ($\pm$1.06)  & 19.80 ($\pm$0.19) \\
Offline re-training + task oracle &  80.42 ($\pm$0.95)& 52.30 ($\pm$0.02) \\
\midrule
\textbf{Single-Head and Task-Agnostic} & & \\
\midrule
\rowcolor{Gray}
HVCL (ours) & 78.41 ($\pm$1.18) & 33.10 ($\pm$0.62) \\
HVCL w/ GR (ours) & 81.00 ($\pm$1.15) & 37.20 ($\pm$0.52) \\
\rowcolor{Gray}
CL-DR \citep{han2021continual} & 86.72 ($\pm$0.30) & 25.62 ($\pm$0.22) \\
NCL \citep{kao2021natural} & & 38.79 ($\pm$0.24) \\ 
\rowcolor{Gray}
TLR \citep{mazur2021target} & 74.89 ($\pm$0.61 & \\
MAS \citep{he2022online} & 73.50 ($\pm$1.54)  & \\
\midrule
\textbf{Multi-Head and Task-Aware} & & \\
\midrule
\rowcolor{Gray}
GEM \citep{lopez2017gradient} & 79.10 ($\pm$1.60) & 40.60 ($\pm$1.90)  \\
MCCL \citep{kj2020meta} & 82.90 ($\pm$1.20) & 43.50 ($\pm$0.60)\\
\rowcolor{Gray}
CCLwFP \citep{han2021contrastive} & 86.33 ($\pm$1.47) & 65.19 ($\pm$0.65)\\
HAL \citep{chaudhry2021using} & 75.19 ($\pm$2.57) & 47.88 ($\pm$2.76)  \\
\rowcolor{Gray}
SI \citep{zenke2017continual} & 63.31 ($\pm$3.79) & 36.33 ($\pm$4.23)\\
AGEM \citep{chaudhry2018efficient} & 74.07 ($\pm$0.76) & 46.88 ($\pm$1.81)\\
\bottomrule
\end{tabularx}
\end{minipage}
\caption{Results in the supervised CL benchmarks. Results were averaged over ten random seeds with the standard deviation given in the parenthesis. We report results of other methods as given in their original studies.}
\label{tab:mnistresults}
\end{table}
We evaluate our approach in current supervised CL benchmarks in Section \ref{sec:scenarios}, in a generative learning setting in Section \ref{subsec:vae}, and in the CRL setup in Section \ref{sec:crl}. We give experimental details in Appendix \ref{app:cifarexperiments}.

\subsection{Continual Supervised Learning Scenarios}
\label{sec:scenarios}

The basic setting of continual learning is defined as an agent which sequentially observes data from a series of tasks while maintaining performance on older tasks. We evaluate the performance of our method in this setting in split MNIST, permuted MNIST, split CIFAR-10/100 (see Table \ref{tab:mnistresults}). We follow the domain incremental setup \citep{van2020brain}, where task information is not available, but we also compare against task-incremental methods, where the task information is available, to give a complete overview of current methods. 

The first benchmark builds on the MNIST dataset. Five binary classification tasks from the MNIST dataset arrive in sequence and at time step $t$ the performance is measured as the average classification accuracy on all tasks up to task $t$. In permuted MNIST the task received at each time step $t$ consists of labeled MNIST images whose pixels have undergone a fixed random permutation. The second benchmark is a variation of the CIFAR-10/100 datasets. In Split CIFAR-10, we divide the ten classes into five binary classification tasks. CIFAR-100 is like the CIFAR-10, except it has 100 classes and tasks are defined as a 10-way classification problem, thus forming ten tasks in total. 

We achieve comparable results to current state-of-the-art approaches (see Table~\ref{tab:mnistresults}) on all supervised learning benchmarks. 
\subsection{Generative Continual Learning}
\label{subsec:vae}
Generative CL is a simple but powerful paradigm \citep{van2020brain}. The main idea is to learn the data generating distribution and simulate data of previous tasks. We can extend our approach to the generative setting by modeling a variational autoencoder using the novel layers we propose in this work. 

We model the distribution of the latent variable $z$ in the variational autoencoder by using a densely connected MoVE layer with $3$ experts. Using multiple experts enables us to capture a richer class of distributions than a single Gaussian distribution could, as is usually the case in simple VAEs.  We can interpret this as $z$ following a Gaussian Mixture Model, whose components are mutually exclusive and modeled by experts. We integrate the generated data by optimizing a mixture of the loss on the new task data and the loss of the generated data We were able to improve our results in the supervised settings by incorporating a generative component, as we show in Table~\ref{tab:mnistresults}. We show additional empirical results in Appendix \ref{app:generative}.

\subsection{Continual Reinforcement Learning}
\label{sec:crl}
\begin{figure*}[t!]
\begin{minipage}{\textwidth}
\centering
\includegraphics[width=0.99\textwidth, trim={7.5cm 2cm 6cm 3.5cm}, clip]{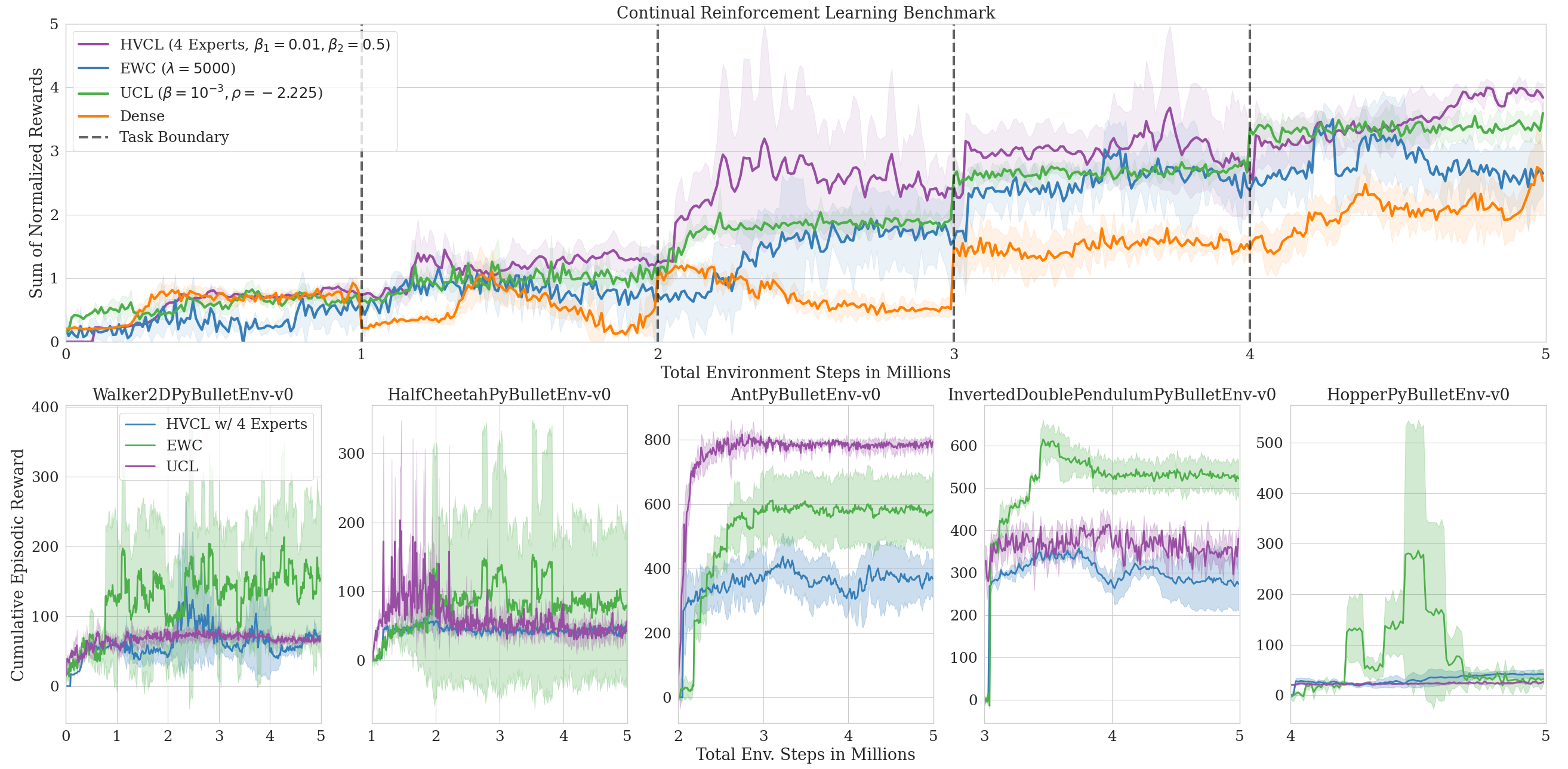}
\end{minipage}
\caption{In this figure we show results in the CRL benchmark. To measure the continual learning performance in the RL settings, we normalize rewards and plot the sum of normalized rewards. The maximum of $5.0$ indicates no forgetting, while $1.0$ shows the total forgetting of old tasks. We comapre against EWC \citep{kirkpatrick2017overcoming}, and UCL \citep{ahn2019uncertainty}.}
\label{fig:crl}
\end{figure*}
In the continual reinforcement learning (CRL) setting, the agent is tasked with finding an optimal policy in sequentially arriving RL problems. To benchmark our method, we follow the experimental protocol of \cite{ahn2019uncertainty} and use a series of tasks from the PyBullet environments \citep{benelot2018pybulletgym}. The environments we selected have different states and action dimensions. This implies we can't use a single model to learn policies and value functions. To remedy this, we pad each state and action with zeros to have equal dimensions. The Ant environment has the highest dimensionality with a state dimensionality of 28 and an action dimensionality of 8. All others are zero-padded to have this dimensionality. 

Here, we extend SAC \citep{haarnoja2018soft} by implementing all neural networks with MovE layers. When a new task arrives, the old posterior over the expert parameters and the gating posterior become the new priors. After each update step in task $t$, we evaluate the agent in all previous tasks $T_{1:t}$ for three episodes each. We divide the reward achieved during evaluation by the mean reward during training and report the cumulative normalized reward, which gives an upper bound of $t$ in the $t$-th task.

We compare our approach against a simple continuously trained SAC implementation with dense neural networks, EWC \citep{kirkpatrick2017overcoming}, and the recently published UCL \citep{ahn2019uncertainty} method. UCL is similar to our approach in that it also employs Bayesian neural networks, but the weight regularization acts on a per-weight basis. Note that UCL and EWC both require task information to compute task-specific losses. Our results (see Figure~\ref{fig:crl}) show that our approach can sequentially learn new policies while maintaining an acceptable performance on previously seen tasks. Our method outperforms UCL \citep{ahn2019uncertainty} and EWC \citep{kirkpatrick2017overcoming}. In this setting naively training the agent sequentially (labeled "Dense") yields poor performance. This behavior indicates the complete forgetting of old policies.

\section{Discussion}
\label{sec:discussion}
The principle we propose in this work falls into a wider class of methods that deal with learning and decision-making problems by integrating information-theoretic cost functions. Such information-constrained machine learning methods have enjoyed recent interest in a variety of research fields, e.g.,\ as reinforcement learning \citep{eysenbach2018diversity,ghosh2018divide,leibfried2019mutual,hihn2019information,arumugam2021information}, MCMC optimization \citep{hihn2018bounded,pang2020learning}, meta-learning \citep{rothfuss2018promp,hihn2020hierarchical}, continual learning \citep{nguyen2017variational,ahn2019uncertainty}, and self-supervised learning \citep{thiam2021multi,tsai2021self}. 

Currently, there are only few methods that perform well in supervised CL and in CRL \cite[e.g.,][]{ahn2019uncertainty,jung2020continual,cha2020cpr}. These methods require task information, as they either keep a set of separate task-specific heads \citep{ahn2019uncertainty,jung2020continual} or compute task-specific losses \citep{cha2020cpr}. This makes our method one of the first task-agnostic CL approaches to do so.

The hierarchical structure we employ is a variant of the Mixture-of-Experts (MoE) model \citep{Jacobs1991}, specifically an extension of the sparsely-gated MoE layers \citep{Shazeer2017}. Sparsely-gated MoE layers enforce a balanced load between experts. In our work, we removed the incentive to equally distribute inputs, as we aim to find specialized experts, which contradicts a balanced load. The computational advantage remains, as we still activate only the top-$1$ expert.

Our method is similar to the approach described by \cite{hihn2020specialization} but differs in two key aspects. Firstly, we provide a more stable learning procedure as our layers can readily offer end-to-end training. Secondly, we implement the information-processing constraints on the parameters instead of the output of the experts, thus shifting the information cost from decision-making to learning. 

Several methods in the current CL literature rest on modular architectures \citep[e.g.,][]{fernando2017pathnet,collier2020routing,lin2019conditional,Lee2020A}. \cite{lin2019conditional} propose to condition model parameters on the inputs  by learning a (deterministic) grouping function. Our approach differs in two main ways. First, our method can capture uncertainty allowing us to learn stochastic tasks. Second, our design can incorporate up to $2^n$ paths (or groupings) through a neural net with $n$ layers, making it more flexible than learning a mapping function. \cite{Lee2020A} propose a MoE model for continual learning, in which the number of experts increases dynamically, utilizing Dirichlet-Process-Mixtures \citep{antoniak1974mixtures} to infer the number of experts. The authors argue that since the gating mechanism is itself a classifier, training it in an online fashion would result in catastrophic forgetting. To remedy this, they implement a generative model per expert $m$ to model $p(m\vert x)$ and approximate the output as $p(y\vert x) \approx \sum_m p(y\vert x)p(m\vert x)$. In our work, we have demonstrated that it is possible to implement a gating mechanism based only on the input by coupling it with an information-theoretic objective to prevent catastrophic forgetting.

Recently, several diversity measures have been proposed. \cite{parker2020effective} introduce a DDP-based method based on the different states a given policy may reach. \cite{dai2021diversity} propose to augment the sampling process in hindsight experience replay \citep{andrychowicz2017hindsight} with a DPP-diversity bonus. The method we propose differs from previous methods as we define diversity in parameter space instead of the policy outcomes or inputs. Additionally, as we define it on parameters instead of actions, we can apply it straightforwardly to any problem formulation, as our experiments show.

\section{Conclusion}
\label{sec:conclusion}
We introduced a hierarchical approach to task-agnostic continual learning, derived an application, and extensively evaluated this method in supervised CL and CRL. While we removed the task-information limitation, we achieved results competitive to task-aware and to task-agnostic algorithms. We argued that both VCL and hierarchical VCL have strong connections to an information-theoretic formulation of bounded rationality. We designed a diversity objective that stabilizes learning and further reduces the risk of catastrophic forgetting. Our method builds on generic utility functions, we can apply it independently of the underlying problem, which makes our method one of the first to do so.

\section*{Acknowledgment}
This work was supported by the European Research Council, grant number ERC-StG-2015-ERC, Project
ID: 678082, ``BRISC: Bounded Rationality in Sensorimotor Coordination".
\bibliographystyle{iclr2022_conference}
\bibliography{merged_bibliography.bib}

\appendix

\section{Wasserstein-Distance between two Gaussians}
\label{app:wassergaussian}
The $W_2^2$ distance between two Gaussians is given by 
\begin{equation}
W_2^2(p,q) = ||\mu_p - \mu_q||_2^2 + ||\sqrt{d_p} - \sqrt{d_q}||_2,
\end{equation}

\textit{Proof:} Let $p = \mathcal{N}(\mu_p, \Sigma_p)$ and $q = \mathcal{N}(\mu_q, \sigma_q)$ be two Guassian distributions. The Wasserstein-2 distance between $p$ and $q$ is then given by
\begin{equation}
W_2^2(p,q) = ||\mu_p - \mu_q||_2^2 + \mathcal{B}(\Sigma_p,\Sigma_q),
\end{equation}
where $\mathcal{B}$ is the Bures metric between two positive semi-definite matrices:
\begin{equation}
\mathcal{B}(\Sigma_p,\Sigma_q) = tr(\Sigma_p + \Sigma_q - 2(\Sigma_p^{1/2} \Sigma_q \Sigma_p^{1/2})^{1/2},
\end{equation}
where $tr(A)$ is the trace of a matrix $A$ and $A^{1/2}$ is the matrix square root. Matrix square roots are computationally expensive to compute and there can potentially be and infinite number of solutions. In the case where $p$ and $q$ are Gaussian mean-field approximations, i.e., all dimensions are independent, $\Sigma_p$ and $\Sigma_q$ are given by diagonal matrices, such that $\Sigma_p = \text{diag}(d_p)_i$ and $\Sigma_q = \text{diag}(d_q)_i$. The Bures metric then reduces to the Hellinger distance between the diagonals $d_p$ and $d_q$, and we have:
\begin{equation}
W_2^2(p,q) = ||\mu_p - \mu_q||_2^2 + ||\sqrt{d_p} - \sqrt{d_q}||_2.
\label{eq:wassernorm}
\end{equation}

\section{Wasserstein-2 Exponential Kernel}
\label{app:wasserkernel}
The exponential Wasserstein-2 kernel between isotropic Gaussian distributions $p$ and $q$ with kernel width $h$ defined by
\begin{equation*}
k(p,q) = \exp\left(-\frac{W_2^2(p,q)}{2h^2}\right)
\end{equation*} is a valid kernel function.

\textit{Proof:} The simplest way to show a kernel function $k$ is valid is by deriving $k$ from other valid kernels. We can express the Wasserstein distance as the sum of two norms as shown in \eqref{eq:wassernorm}. The euclidean norm and the Hellinger distance both form inner product spaces and are thus valid kernel functions. Their sum is also a valid kernel function, which makes the Wasserstein distance on isotropic Gaussians a valid kernel. If $k(p,q)$ is a valid kernel, then $\exp(k(p,q))$ is also a valid kernel.

\section{Experiment Details}
\label{app:cifarexperiments}
To implement variational layers we use Gaussian distributions. For simplicity we use a $D-$dimensional Gaussian mean-field approximate posterior $q_t(\theta)= \prod^D_{d=1}\mathcal{N}(\theta_t\vert p_{t,d},\sigma^2_{t,d})$. We use the flip-out estimator \citep{wen2018flipout} to approximate the gradients. In practice, we draw a single sample to approximate the expectation. 
\subsection{MNIST Experiments}
For split MNIST experiments we used dense layers for both the VAE and the classifier. The VAE encoder contains two layers with 256 units each, followed by 64 units (64 units for the mean and 64 units for log-variance) for the latent variable, and two layers with 256 units for the decoder, followed by an output layer with $28*28 = 784$ units. This assumes isotropic Gaussians as priors and posteriors over the latent variable and allows to compute the $\DKL$ if closed form. We used only one expert for the VAE with $\beta_1 = 0.002$, $\beta_2 = 0.75$, a diversity bonus weight of $0.01$ and  leaky ReLU activations \citep{maas2013rectifier} in the hidden layers. We trained with a batch size 256 for 150 epochs. The VAE output activation function is a sigmoid and we trained it using a binary cross-entropy loss between the normalized pixel values of the original and the reconstructed images.  We used no other regularization methods on the VAE. We used 10.000 generated samples after each task.

The classifier consists of two dense layers, each with 256 units with leaky ReLU activations \citep{maas2013rectifier} and dropout \citep{srivastava2014dropout} layers, followed by an output layer with two units. All layers of the classifier have two experts. We trained with batch size 256 for 150 epochs using Adam \citep{KingmaBa2015} with a learning rate of $6*10^{-4}$. In the permuted MNIST setting we used the same architecture, but increased the number of units to 512.

\subsection{CIFAR-10 Experiments}
The VAE encoder consisted of five convolutional layers with stride 4 with two experts, each with 8, 16, 32, 64, and 128 units, followed by two dense units with two experts, each with 256 units. The latent variable has 128 dimensions, which we model by two dense layers: one with 128 units for the mean and one with 128 units for the log-variance. The dense layer modeling the mean has three experts, the layer for the log-variance one expert. We assume isotropic Gaussian distributions as priors and posteriors over the latent variable, which allows us to compute the $\DKL$ if closed form. The decoder mirrors the encoder and has two dense layers followed by 5 de-convolutional layers with stride 4  (the last layer has stride 3). All hidden layers use a leaky ReLU activation function \citep{maas2013rectifier}. The VAE output activation function is a sigmoid and we trained it using a binary cross-entropy loss between the normalized pixel values of the original and the reconstructed images. We used no other regularization methods on the VAE. We used 10.000 generated samples after each task.

The classifier architecture is similar to the encoder architecture. We used five convolutional layers, followed by two dense layers. All layers used two experts. The convolutional layers have 8, 16, 32, 64, and 128 units per experts, while the dense layers both have 256 units per layer. We used leaky ReLU as an activation function for the hidden layers and softmax for the output layer. We trained the classifier using a binary cross-entropy loss between the true and the predicted label. We trained with batch size 256 for 1000 epochs using the Adam optimizer with a learning rate of $3*10^{-4}$.

\subsection{Reinforcement Learning Experiment Details}
\label{app:crlexperiments}
Each task was trained for one million time steps. We use the same network architecture as suggested by the authors UCL: two layer networks (actor and critics) with 16 units each. Each layer has four experts followed by leaky ReLU \citep{maas2013rectifier} activation functions. Each We set each SAC related hyper-parameter as proposed in the original publication \citep{haarnoja2018soft}.
For UCL \citep{ahn2019uncertainty}, we used the implementation provided by the authors for our experiments and use the hyper-parameters suggested in the publication. Note that the UCL implementation rests on a PPO \citep{Schulman2017} backbone. Our CRL experiments do not use any form of replay (except for the replay buffer used by SAC).

\section{Additional Experiments}
\subsection{Generative CL}
\label{app:generative}
 \begin{figure}[t!]
\centering
\begin{minipage}{0.66\columnwidth}
\includegraphics[width=\textwidth, trim={2.25cm 0.25cm 3cm 1cm}, clip]{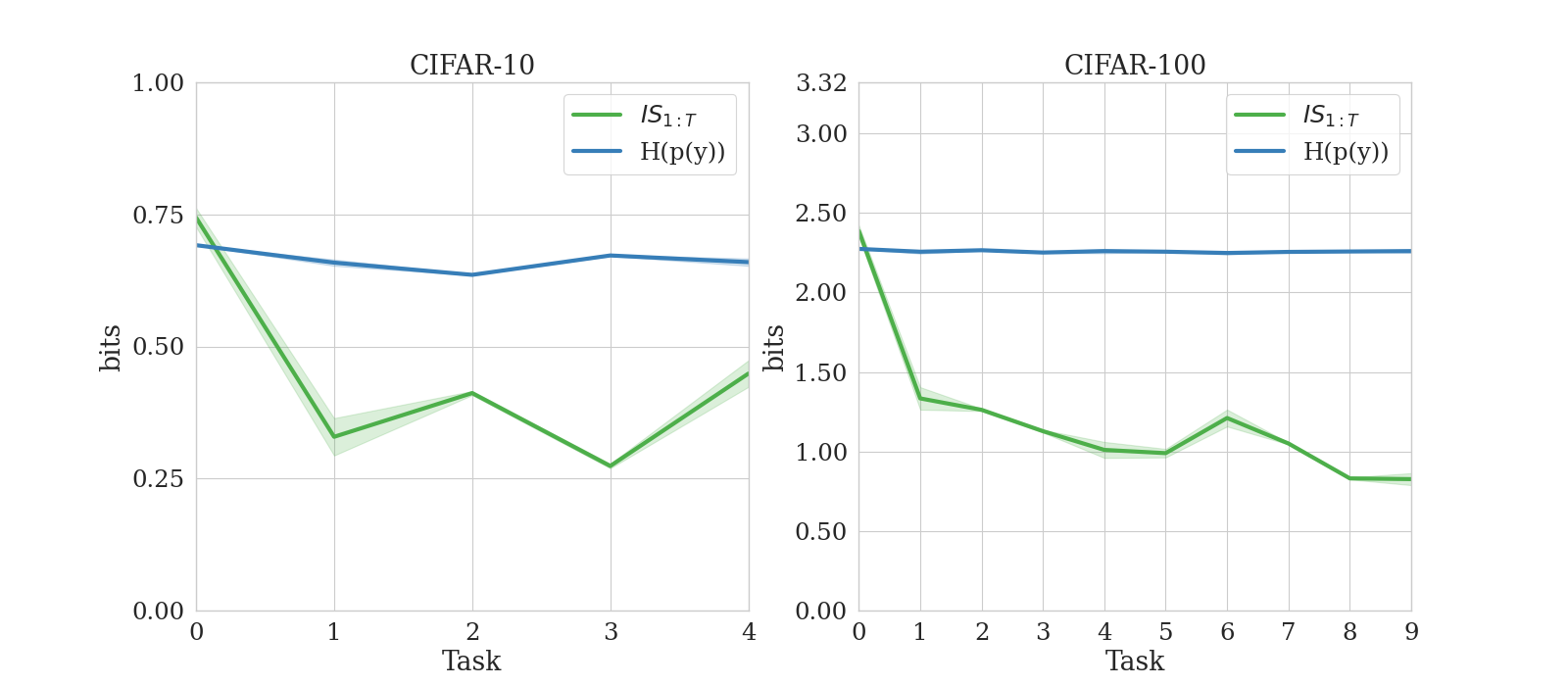}
\end{minipage} 
\begin{minipage}{0.32\columnwidth}
\includegraphics[width=\textwidth, trim={0cm 2.15cm 0cm 0cm}, clip]{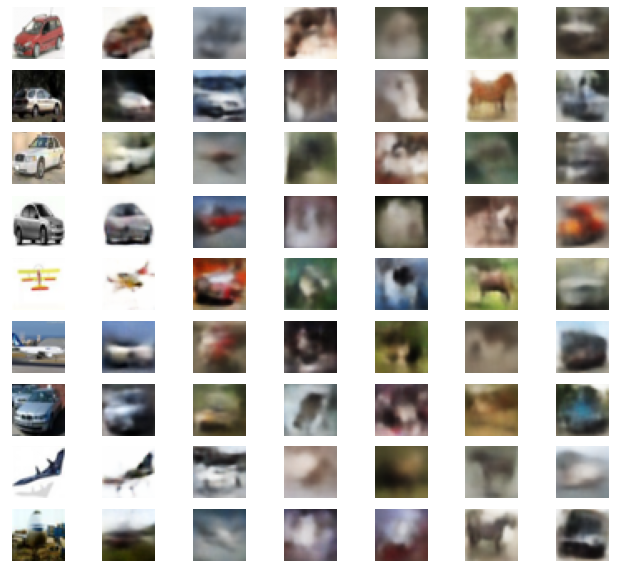}
\end{minipage} 
\caption{Left and middle: measures for the generator quality. Right: The first two rows depict original images and the corresponding reconstructions. The following five rows are images sampled from the VAE prior after each task (automobiles vs.\ airplanes, birds vs.\ cats, deers vs.\ frogs, dogs vs.\ horses, and ships vs.\ trucks).}
\label{fig:genquality}
\end{figure}
We introduce a generative approach to continual learning in Section \ref{subsec:vae} by implementing a Variational Auto-encoder using our proposed layer design. This addition improved classification performance and mitigated catastrophic forgetting, as evidenced by the results shown in Table \ref{tab:mnistresults}. We train by integrating artificially generated data into the process by optimizing a mixture loss:
\begin{equation}
L(\theta) = \frac{1}{2\vert B_{t}\vert } \sum_{b\in B_t}\ell(b) + \frac{1}{2\vert B_{1:t}\vert } \sum_{b\in \vert B_{1:t}\vert }\ell(b),
\end{equation}
where $B_t$ is batch of data from the current task, $B_{1:t}$ a batch of generated data, and $\ell(b)$ a loss function on the batch $b$.

Borrowing methods from generative learning, we investigate the performance of our proposed VAE design further, with the main focus on the quality of the generated images. We can not measure the accuracy directly, as artificial images lack labels. Thus we first use the trained classifier to obtain labels and compute metrics based on these self-generated labels. We opted for the Inception Score (IS) \citep{salimans2016improved}, as it is widely used in the generative learning community. In this initially proposed formulation, the IS builds on the $\DKL$ between the conditional and the marginal class probabilities as returned by a pre-trained Inception model \citep{szegedy2015going}. To investigate the quality of the generated images concerning the continually trained classifier, we use a different version of the Inception Score, which we defined as
\begin{equation}
IS_{T}(G_{1:T}) = \mathbb{E}_{x\thicksim G_{1:T}}\left[ \DKL\left[p_{1:T}(y\vert x) \vert \vert  p_{1:T}(y)\right] \right],
\end{equation}
where $G_{1:T}$ is the data generator trained on tasks up to $T$, $p_{1:T}(y\vert x)$ the conditional class distribution returned by the classifier trained up to task $T$, and $p_{1:T}(y)$ the marginal class distribution up to Task $T$. Note that, $IS(G_{1:T}) \leq \log_2 N_c$, where $N_c$ is the number of classes. We show $IS_{T}$ and the entropy of $p(y)$ in the split CIFAR-10 and split CIFAR-100 setting in Figure~\ref{fig:genquality}.


\end{document}